\documentclass[letterpaper, 10 pt, conference]{ieeeconf}  
\usepackage{cite}
\usepackage{amsmath,amssymb,amsfonts}
\usepackage{graphicx}
\usepackage{textcomp}
\usepackage{xcolor}
\usepackage{multirow}
\usepackage{booktabs}
\usepackage{subcaption}
\usepackage{balance}
\usepackage{booktabs}        
\usepackage{algorithm}
\usepackage{algpseudocode}
\usepackage{threeparttable}
\usepackage[draft=true]{hyperref}
\IEEEoverridecommandlockouts                              

\title{\LARGE \bf
Uncovering Critical Features for Deepfake
Detection through the Lottery Ticket Hypothesis
}

\author{Lisan Al Amin$^{1,*}$, Md. Ismail Hossain$^{2,*}$, Thanh Thi Nguyen$^{3}$, Tasnim Jahan$^{4}$, Mahbubul Islam$^{4}$, Faisal Quader$^{5}$%
\thanks{$^{*}$Equal contribution}%
\thanks{$^{1}$Cognitive Links, Maryland, USA}%
\thanks{$^{2}$Apurba-NSU R\&D Lab, North South University, Bangladesh}%
\thanks{$^{3}$AiLECS Lab, Monash University, Australia}%
\thanks{$^{4}$United International University, Bangladesh}%
\thanks{$^{5}$University of Maryland, Baltimore County, USA}%
}

\begin{document}

\maketitle
\thispagestyle{empty}
\pagestyle{empty}

\begin{abstract}

Recent advances in deepfake technology have created increasingly convincing synthetic media that poses significant challenges to information integrity and social trust. While current detection methods show promise, their underlying mechanisms remain poorly understood, and the large sizes of their models make them challenging to deploy in resource-limited environments. This study investigates the application of the Lottery Ticket Hypothesis (LTH) to deepfake detection, aiming to identify the key features crucial for recognizing deepfakes. We examine how neural networks can be efficiently pruned while maintaining high detection accuracy. Through extensive experiments with MesoNet, CNN-5, and ResNet-18 architectures on the OpenForensic and FaceForensics++ datasets, we find that deepfake detection networks contain winning tickets, i.e., subnetworks, that preserve performance even at substantial sparsity levels. Our results indicate that MesoNet retains 56.2\% accuracy at 80\% sparsity on the OpenForensic dataset, with only 3,000 parameters, which is about 90\% of its baseline accuracy (62.6\%). The results also show that our proposed LTH-based iterative magnitude pruning approach consistently outperforms one-shot pruning methods. Using Grad-CAM visualization, we analyze how pruned networks maintain their focus on critical facial regions for deepfake detection. Additionally, we demonstrate the transferability of winning tickets across datasets, suggesting potential for efficient, deployable deepfake detection systems.

\end{abstract}

\section{Introduction}
The exponential growth in generative AI capabilities has revolutionized digital content creation, enabling the production of synthetic media that is increasingly indistinguishable from authentic content. Deepfake technology enables convincing synthetic media, raising concerns about information integrity, social trust, and security \cite{shoaib2023deepfakes, pramod2024generative}. Current deepfake detection methods, e.g., in \cite{mao2023deepfake,hu2021detecting,sabareshwar2024lightweight}, though promising, often operate as black boxes, making it difficult to understand their decision-making processes or guarantee their reliability and feasibility across diverse scenarios, particularly when deployed in complex and resource-constrained environments. This gap in understanding deepfake detection vs. traditional vision tasks presents a key research opportunity. This research aims to address several critical questions: (1) How do feature learning mechanisms in deepfake detection differ from traditional image classification? (2) Can analyzing model pruning provide insights into the essential features for detecting synthetic content? and (3) How can understanding these differences inform the development of more efficient detection systems?

\begin{figure}[t]
    \centering
    \includegraphics[width=0.8\linewidth]{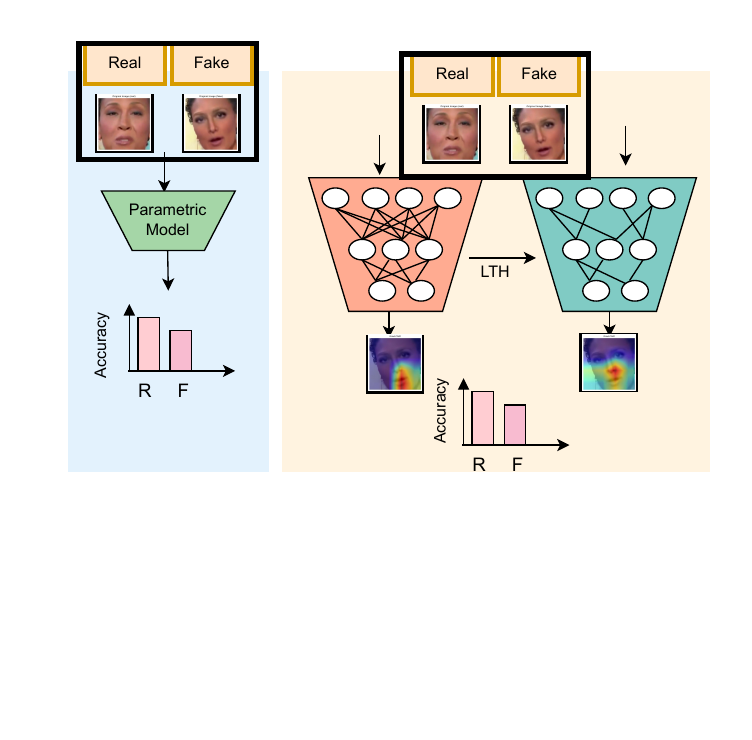}
    \vspace*{-21 mm}
    \caption{Comparison of deepfake detection using a dense model (left) and a pruned model (right) with Grad-CAM visualizations. The pruned model retains performance despite fewer parameters.}
    \label{fig:deepfake_comparison}
\end{figure}

Previous research in neural network compression has revealed that traditional image classification models contain significant redundancy, with studies demonstrating that up to 80\% of weights can be pruned while maintaining acceptable performance\cite{frankle2018lottery}. This suggests that these models primarily rely on a small subset of weights to learn essential features. However, deepfake detection presents unique challenges that may not follow this pattern. While existing studies have explored various detection approaches, from analyzing visual artifacts to leveraging temporal inconsistencies \cite{gu2023exploiting}, the underlying feature learning mechanisms remain poorly understood. The limited investigation into how pruning affects deepfake detection models specifically presents an opportunity to advance our understanding of these critical systems.

This study proposes a comprehensive comparative analysis framework for deepfake detection using iterative magnitude-based weight pruning, particularly the Lottery Ticket Hypothesis (LTH), which posits that randomly initialized neural networks contain subnetworks (termed ``winning tickets'') that can independently achieve performance comparable to the original network after sufficient training \cite{frankle2018lottery}. The research methodology begins with gradual weight removal based on magnitude values, systematically pruning neural network weights. It then conducts a thorough analysis of model behavior at different pruning thresholds to understand how weight removal affects performance. The methodology also performs a detailed examination of persistent features through progressive pruning, which includes visualization of Grad-CAM under various compression rates. The ultimate goal is to bridge the gap between theoretical understanding and practical application, potentially leading to more efficient and accessible deepfake detection systems while providing valuable insights into how these models differ fundamentally from traditional image classification tasks. 

Our extensive experiments with MesoNet\cite{afchar2018mesonet}, CNN-5, ResNet-18 architectures \cite{he2016deep}, and XceptionNet \cite{Rssler2018FaceForensicsAL} on the OpenForensic\cite{Kuznetsova-IJCV2020} and FaceForensics++ \cite{rossler2019faceforensics++} datasets reveal several crucial insights into deepfake detection. Through Iterative Magnitude-Based Pruning (IMP), we discovered that deepfake detection models exhibit unique sparsity characteristics. Although MesoNet maintains accuracy at 80\% sparsity in the OpenForensic dataset, its performance drops significantly beyond this threshold, suggesting a minimum critical feature retention requirement. CNN-5 demonstrated similar behavior on FaceForensics++, achieving 78.3\% accuracy at 60\% sparsity but showing instability with further pruning. Notably, ResNet-18 proved the most resilient, maintaining accuracy even at extreme sparsity, indicating its superior feature extraction capabilities. Our Grad-CAM analysis revealed that pruned models increasingly focus on high-frequency artifacts and temporal inconsistencies rather than semantic features. Through this process, we aim to uncover the critical features required for detecting synthetic content and analyze how these differ from traditional image classification tasks. In summary, our contributions are as follows:

\begin{itemize}

    \item  We provide comprehensive sparsity analysis of different deepfake detection architectures, revealing unique compression characteristics compared to traditional vision models.
    
    \item We demonstrate through systematic experiments that different architectures exhibit varying resilience to pruning, with ResNet-18 showing superior feature retention at extreme sparsity levels, up to 80\%, without compromising accuracy.

    \item We offer insights through Grad-CAM analysis, showing how pruned models shift focus to high-frequency artifacts rather than semantic features, advancing our understanding of essential features for synthetic content detection.
\end{itemize}

\section{Literature Review}
\subsection{Deepfake Detection Models and Evolution}
Several methods have been developed to improve the detection of compressed deepfake content. One such prominent strategy is to combine dense and dynamic CNNs with sophisticated algorithms, such as face clustering and attention mechanisms \cite{mao2023deepfake}. Similarly, a two-stream convolutional network that uses frame-temporality can improve identification accuracy by leveraging temporal irregularities in compressed videos \cite{hu2021detecting}. Lightweight CNNs in the frequency domain have been developed for low-resolution material, efficiently discovering minor discrepancies that classic spatial domain-based approaches frequently ignore~\cite{sabareshwar2024lightweight}. Another important advancement in real-time applications is FakeBuster, a streamlined 3D CNN tool meticulously built to detect deepfakes in video conferences such as Zoom and Skype~\cite{hubens2021fake}.

\subsection{Pruning Algorithms and Their Applications} 
To optimize deep learning models, pruning strategies are crucial, allowing efficiency without sacrificing performance. NGLOD renders high-fidelity neural signed distance functions by pruning within octree-based feature volumes, allowing virtual reality applications and interactive graphics \cite{takikawa2021neural}. The LLM-Pruner approach selectively eliminates less important components from large language models (LLMs) using gradient information by retaining the model's fundamental functionality \cite{ma2023llm}. Furthermore, the work in \cite{yang2024pruning} presents a targeted pruning method designed to selectively eliminate particular concepts from diffusion models, ensuring the preservation of overall model performance. Pruning generative models, such as StyleGAN2-ADA, improve computational performance in video editing, virtual reality, and recreation, making generative AI more accessible \cite{takikawa2021neural}. According to the work in \cite{molchanov2017pruning}, pruned networks can be used in medical imaging to identify high-accuracy tumors on portable devices, allowing healthcare providers to perform diagnostic activities seamlessly. Pruning techniques are also employed in voice recognition, where structured RNN pruning enables real-time transcription and portable virtual assistants \cite{dong2020rtmobile}.
 All these algorithms have an abundant contribution in the development of deepfake detection domain, but leaving the interpretability of deepfake models unexplored. Understanding the characteristics learned by deepfake models is critical for enhancing detection systems and enabling the selective pruning of deepfake models.

\section{Exploring the Lottery Ticket Hypothesis for Deepfake Detection}


The LTH suggests that within a randomly initialized neural network, there exists a smaller, more efficient sub-network (the ``lottery ticket'' or ``winning ticket'') that, if trained in isolation, can perform just as well as the original, larger network. In the domain of deepfake detection, this study focuses on identifying and evaluating these winning tickets within a neural network, \(\mathcal{F}(\mathbf{x}; \theta)\), which maps input images \(\mathbf{x} \in \mathcal{X}\) to binary labels \(\mathcal{Y}\). The neural network parameters \(\theta\) are initialized from a distribution \(\mathcal{D}_\theta\).

\subsection{ Identifying Winning Tickets}
To extract the winning ticket subnetwork, we implement an iterative process of pruning and retraining, as described below~\cite{Biswas_2024_BMVC}:

\begin{itemize}\setlength{\itemsep}{-0.2em}
    \item[1.] Initialize a binary mask: Let us denote the mask as \(\mathbf{m} \in \{0, 1\}^{|\theta|}\), with all elements set to one: \(\mathbf{m} = \mathbf{1}\).
    \item[2.] Train neural network: Apply the training process to the network \(\mathcal{F}(\mathbf{x}; \mathbf{m} \odot \theta)\) for a single cycle.
    \item[3.] Prune weights: Remove a percentage \(p\) of the weights by updating the mask \(\mathbf{m}\). Specifically, set \(\mathbf{m}_i = 0\) if \(|\theta_i| \leq \alpha\); otherwise, retain \(\mathbf{m}_i = 1\), where \(\alpha\) is a threshold.
    \item[4.] Iterate: Rerun steps 2 and 3 for \(j\) training cycles or until the expected sparsity level is reached.
\end{itemize}

The goal is to discover a sparse subnetwork \(\mathcal{F}(\mathbf{x}; \mathbf{m} \odot \theta)\) that achieves accuracy \(a' \geq a\), where \(a\) is the accuracy of the original dense network, after \(J\) pruning iterations.

\subsection{Challenges in Applying LTH to Deepfake Detection}
The application of LTH to deepfake detection introduces specific challenges related to feature preservation and pruning biases:

\subsubsection{Preserving Rare Feature Sensitivity}
Deepfake detection critically depends on recognizing subtle manipulation artifacts \cite{kaur2024deepfake}, often represented by a limited number of parameters in the network. Let \(\mathcal{R}\) denote the set of rare but essential features \(\{r_1, \ldots, r_k\}\) relevant to detecting deepfakes. The winning ticket subnetwork \(\mathcal{F}(\mathbf{x}; \mathbf{m} \odot \theta)\) must retain sensitivity to these features:
\[
P(\mathcal{F}(\mathbf{x}; \mathbf{m} \odot \theta) | r_i) \geq \tau, \quad \forall r_i \in \mathcal{R},
\]
where \(\tau\) is the minimum required sensitivity threshold.

\subsubsection{Mitigating Pruning Bias}
Magnitude-based pruning methods tend to remove weights linked to infrequent features~\cite{hooker2019compressed}, which may obstruct the identification of rare artifacts. Let \(f_i\) represent the frequency of feature \(i\) in the training data. The likelihood of retaining weights \(\theta_i\) linked to feature \(i\) is influenced by:
\[
P(\mathbf{m}_i = 1) \propto f_i \cdot \|\theta_i\|.
\]
This introduces a systematic bias against rare but critical features necessary for deepfake detection.

\begin{algorithm}[ht]
\caption{Iterative Magnitude Pruning (IMP)}
\label{alg:IMP}
\begin{algorithmic}[1]
\State Initialize weights $\theta$ and mask $m \leftarrow 1$
\For{$j = 1, 2, \ldots, J$}
    \State Train $F(x; m \odot \theta)$ for $T$ epochs
    \State Compute global or local pruning threshold $\alpha^{(j)}$
    \For{each layer $l$ and weight index $i$}
        \If{$|\theta_l[i]| \leq \alpha^{(j)}$}
            \State $m_l[i] \leftarrow 0$ \Comment{Prune the weight}
        \EndIf
    \EndFor
    \State $\theta \leftarrow m \odot \theta$
\EndFor
\end{algorithmic}
\end{algorithm}

\subsection{Iterative Magnitude Pruning (IMP) }

To address the challenges mentioned above, we propose an IMP approach with the following steps:

\begin{itemize}
    \item \textbf{Initialization:} Begin with a fully dense neural network \(\mathbf{m}^{(0)} = \mathbf{1}\).
    \item \textbf{Training and Pruning:} For each pruning round \(j = 1, 2, \ldots, J\):
    \begin{itemize}
        \item Train the network \(\mathcal{F}(\mathbf{x}; \mathbf{m}^{(j-1)} \odot \theta^{(j-1)})\) for a single cycle.
        \item Update the mask \(\mathbf{m}^{(j)}\) using global or local pruning thresholds \(\alpha^{(j)}\):
        \begin{itemize}
            \item \textbf{Global pruning:} Define a global threshold \(\alpha^{(j)}\) based on the distribution of all weights in the network. Prune weights across all layers based on their global ranking:
            \[
            \mathbf{m}_l^{(j)}[i] = 
            \begin{cases} 
            0 & \text{if } |\theta_l^{(j-1)}[i]| \leq \alpha^{(j)}, \\ 
            1 & \text{otherwise.} 
            \end{cases}
            \]
            \item \textbf{Local pruning:} Apply a distinct pruning threshold \(\alpha_l^{(j)}\) for each layer \(l\), ensuring layer-wise control over sparsity:
            \[
            \mathbf{m}_l^{(j)}[i] = 
            \begin{cases} 
            0 & \text{if } |\theta_l^{(j-1)}[i]| \leq \alpha_l^{(j)}, \\ 
            1 & \text{otherwise.} 
            \end{cases}
            \]
        \end{itemize}
        \item Update the weights: \(\theta^{(j)} = \mathbf{m}^{(j)} \odot \theta^{(j-1)}\). We summarize the IMP procedure in Algorithm 1 for clarity.
    \end{itemize}
    \item \textbf{One-shot pruning:} Apply the mask update \(\mathbf{m}^{(1)}\) after a single training cycle.
\end{itemize}

As an alternative to iterative pruning, \emph{one-shot pruning} implements the mask update \(\mathbf{m}^{(1)}\) after completing one full training cycle, immediately reducing the network’s size based on the initial distribution of weight magnitudes. This method is computationally efficient but sacrifices the granularity and iterative refinement provided by IMP.

IMP investigates how pruning impacts the network’s ability to retain essential features, particularly rare ones. By analyzing the distribution of pruned weights, we evaluate the frequency and magnitude of features removed at each iteration. This analysis is extended to distinguish global pruning, which optimizes sparsity across the entire network, from local pruning, which ensures critical features in each layer are preserved. Additionally, one-shot pruning provides a fast alternative for rapid prototyping or when computational resources are constrained. However, its limited sensitivity to subtle feature dependencies often results in a trade-off between achieving sparsity and maintaining robust feature detection.

\subsection{Transferability of Sparse Subnetworks Across Datasets}
We evaluate the transferability of winning ticket subnetworks across datasets containing different types of deepfake manipulations. Consider two datasets, \(\mathcal{D}_1\) and \(\mathcal{D}_2\), and their respective networks \(\mathcal{F}_1(\mathbf{x}; \theta_1)\) and \(\mathcal{F}_2(\mathbf{x}; \theta_2)\), both initialized from \(\mathcal{D}_\theta\). After identifying a sparse subnetwork \(\mathcal{F}_1(\mathbf{x}; \mathbf{m}_1 \odot \theta_1)\) from \(\mathcal{F}_1\), the transferability is measured as:
\[
a_2' \approx a_2, \quad \text{where } \theta_2 = \mathbf{m}_1 \odot \theta_1.
\]
Here, \(a_2'\) represents the accuracy of the sparse network on \(\mathcal{D}_2\), and \(a_2\) is the accuracy of the original dense network.

\subsection{Analyzing Pruning Effects Using Grad-CAM}

To assess the impact of pruning on the network’s ability to focus on discriminative regions, we utilize Gradient-weighted Class Activation Mapping (Grad-CAM)~\cite{selvaraju2020grad,nguyen2024adaptive}. Grad-CAM allows us to visualize the regions of the input that the network prioritizes during decision-making, helping to determine whether pruning influences the network's focus on the most relevant features for deepfake detection. By comparing the Grad-CAM heatmaps before and after pruning, we can evaluate how the network's attention shifts and whether it continues to highlight critical areas of the input, ensuring that essential discriminative features are still effectively utilized.

\subsubsection{Feature Activation Analysis}

For a convolutional layer \(l\), let \(A^l\) represent the feature map activations. Grad-CAM computes class-specific importance weights \(\alpha_k^c\) for feature maps using the gradients of the class score \(y^c\) with respect to \(A^l\). These weights highlight the contribution of each feature map to the prediction.

\subsubsection{Heatmap Generation}

Using the importance weights, the Grad-CAM heatmap \(L^c\) is obtained as:
\[
L^c = \text{ReLU}\left(\sum_k \alpha_k^c A^k\right),
\]
where \(\alpha_k^c\) represents the importance of the \(k\)-th feature map. This heatmap emphasizes the regions in the input that influence the prediction, helping visualize the model's focus.

\subsubsection{Comparative Attention Analysis}

To evaluate pruning effects, we compare Grad-CAM heatmaps before and after pruning at different sparsity levels. Let \(L^c_{\text{original}}\) denote the heatmap before pruning and \(L^c_p\) the heatmap at pruning rate \(p\). Changes in attention are quantified as:
\[
\Delta L^c(p) = \|L^c_{\text{original}} - L^c_p\|_F,
\]
where \(\|\cdot\|_F\) is the Frobenius norm. This analysis helps identify whether pruning disrupts the network's ability to focus on critical regions, ensuring essential features are preserved.

\section{Datasets and Experiments}

\noindent \subsection{Datasets}
We utilized two datasets for deepfake detection. A curated dataset from the OpenForensics\cite{Kuznetsova-IJCV2020} dataset, which contains 140,000 images of real and fake samples for training and 11,000 images for testing\footnote[1]{https://www.kaggle.com/datasets/manjilkarki/deepfake-and-real-images}. A combined FaceForensics++\cite{rossler2019faceforensics++} \& Celeb-DF\cite{li2020celeb} dataset\footnote[2]{https://www.kaggle.com/datasets/nanduncs/1000-videos-split}, which comprises 400 videos, equally split between 200 real and 200 fake videos. By extracting 40 face-cropped frames from each video, we generated a total of 11,600 images for training, with 2,400 images used for evaluation.

\subsection{Network Architectures}
We investigate four advanced neural network architectures for deepfake identification: MesoNet \cite{afchar2018mesonet}, CNN-5, ResNet-18 \cite{he2016deep}, and XceptionNet \cite{Rssler2018FaceForensicsAL}. MesoNet, a lightweight neural network, focuses on mesoscopic-level characteristics and manipulation artifacts. The 5-layer CNN provides a simplified convolutional architecture with batch normalization, facilitating rapid training and deployment. ResNet-18 is a relatively large residual-based CNN network that has the ability to extract deep features from data. XceptionNet, built on depthwise separable convolutions, enables highly efficient feature extraction with fewer parameters, and has been widely adopted as a strong baseline for forgery detection tasks due to its ability to capture subtle facial anomalies.

\subsection{Training Protocol}
All experiments utilize PyTorch 1.8.0 framework with the Adam optimizer (learning rate = 1e-4). We implement three pruning strategies: IMP with both global and local criteria, and one-shot pruning as a baseline. The IMP methodology is conducted over 8 iterative rounds, with pruning rates progressively upto 80\%. During each iteration of IMP, we prune 20\% of the weights from the model, applying this pruning uniformly across both convolutional (Conv) layers and multilayer perceptron (MLP) layers to ensure consistent treatment throughout the network. 

Pruning is conducted under two distinct criteria:

\textit{Global Pruning:} We evaluate the importance of weights across the entire network, irrespective of their layer, and prune the least significant 20\% in each iteration.

\textit{Local Pruning:} Each layer is treated independently, pruning 20\% of the least significant weights within each specific layer. 

Training continues until convergence with early stopping (patience = 10) on a single NVIDIA V4090 GPU.

\textbf{Evaluation:} In accordance with a previous study on LTH in \cite{frankle2018lottery}, we evaluate our work by examining testing accuracy and the network pruning rate, expressed as a percentage of sparsity.

\begin{figure*}[t!]
\centering
\begin{center}
\includegraphics[width=1.0\textwidth]{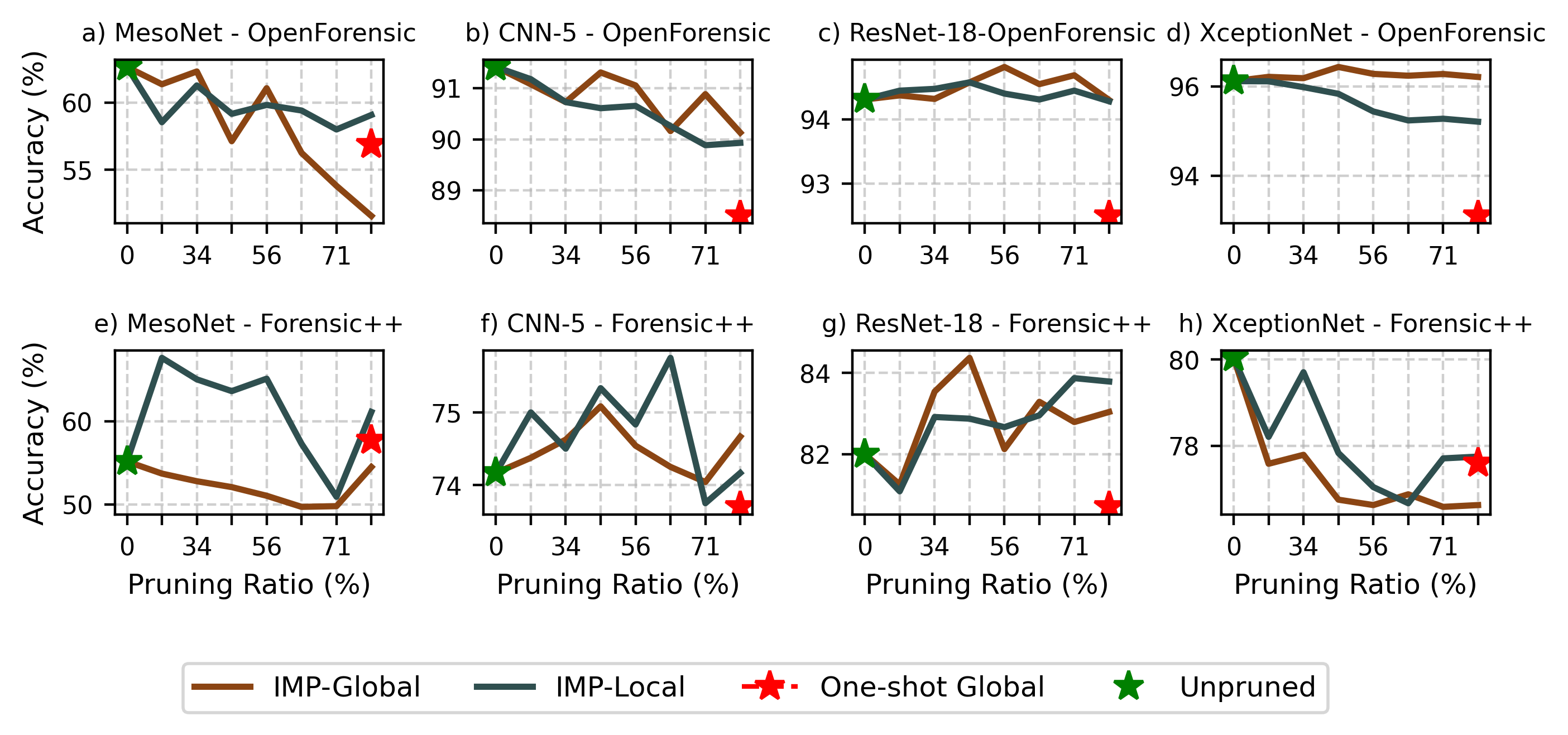}
\end{center}
\vspace{-0em}
\caption{\small The performance of lottery tickets identified for deepfake detection across three different model architectures: MesoNet, CNN-5, ResNet-18 and XceptionNet which utilize distinct approaches for image analysis. Pruning performance varies across datasets. On the OpenForensic dataset, IMP-Global pruning outperforms IMP-Local in 4 out of 4 models. In contrast, on the Forensic++ dataset, IMP-Local consistently surpasses both the overparameterized baseline (0\% pruning) and IMP-Global pruning, demonstrating its task-specific effectiveness. The One-shot Global pruning at 80\% sparsity is also performing adequately. The IMP method for each task progressively prunes 20\% of the weights in each iteration to uncover the winning tickets or subnetworks, delivering competitive or even superior performance relative to overparameterized models, while requiring fewer computations and parameters.}
\label{fig:overall_score}
\end{figure*}

\section{Main Results}

\subsection{Performance Across Pruning Rates}
Fig. \ref{fig:overall_score} presents the comparative analysis across architectures and datasets. On the OpenForensic dataset, IMP-Global pruning consistently outperforms other methods. MesoNet achieves 56.2\% accuracy at 80\% sparsity, which is approximately 89.8\% of its dense model performance (62.6\%). CNN-5 retains 90.88\% accuracy at 80\% sparsity (\(\approx 99.4\%\) of 91.41\%), ResNet-18 retains 94.29\% (\(\approx 99.99\%\) of 94.3\%), and XceptionNet retains 96.11\% (\(\approx 99.9\%\) of 96.20\%). These results confirm that global weight importance generalizes well in this setting.

Conversely, the FaceForensics++ dataset reveals a different trend: IMP-Local pruning proves more effective, especially for deeper models like ResNet-18 , which retains 83\% accuracy at 80\% sparsity under IMP-Local, outperforming both the overparameterized baseline and IMP-Global. MesoNet, CNN-5, and XceptionNet also show improved robustness with IMP-Local compared to other methods.

These findings suggest that pruning strategy effectiveness is dataset-dependent. Global pruning is more stable for cleaner, less compressed data (OpenForensic), whereas local, task-specific pruning better handles challenging, heavily-compressed datasets (FaceForensics++).

\subsection{Impact of Pruning Strategies}
IMP-Local mostly outperforms both IMP-Global and one-shot pruning across all architectures and datasets, which contrasts with regular image classification tasks. The performance gap becomes particularly pronounced at higher sparsity levels ($>$70\%). At 80\% sparsity, IMP-Local maintains a 1-2\% accuracy advantage over one-shot pruning for all models on both datasets. The difference is even more significant on FaceForensics++, with up to 2-3\% improvement in accuracy. This trend reveals that deepfake detection relies more on task-specific features than global generalization, particularly in compressed datasets. Local pruning preserves critical layer-specific patterns, possibly related to subtle compression artifacts, suggesting that uniform global pruning may discard fragile discriminative cues.

\subsection{Architecture-Specific Behavior}
ResNet-18 exhibits the most robust lottery ticket behavior, maintaining near-original performance up to 80\% sparsity on both the dataset. This suggests that deepfake detection features can be effectively captured by a sparse subset of the original network. CNN-5's performance curve shows a more gradual degradation, while MesoNet demonstrates higher sensitivity to pruning. This contrast among architectures highlights an important insight. Larger models like ResNet-18 have more redundancy and thus stronger lottery ticket properties, while compact models like MesoNet are more brittle under compression. This reinforces the idea that subnetwork robustness varies with architectural depth and width.

\subsection{Grad-CAM Analysis}

\begin{figure*}[ht]
\centering
    \begin{subfigure}[b]{0.115\textwidth}
        \centering
        \includegraphics[width=\textwidth]{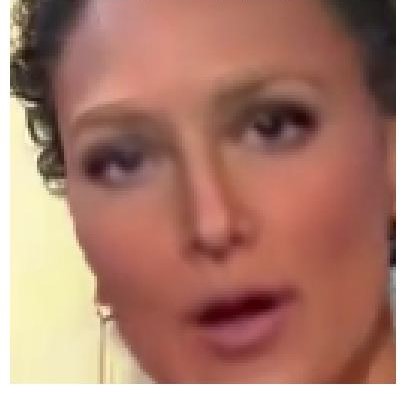}
        \caption*{(a) Fake}
    \end{subfigure}
    \hspace{1pt}
    \begin{subfigure}[b]{0.115\textwidth}
        \centering
        \includegraphics[width=\textwidth]{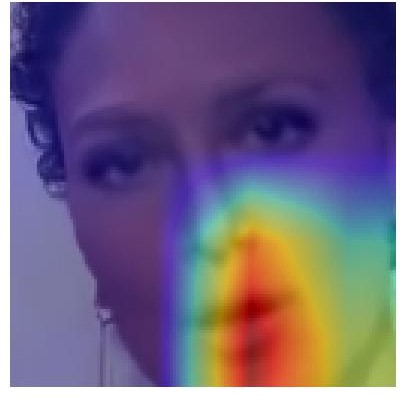}
        \caption*{(b) 0\% Sparse}
    \end{subfigure}
    \hspace{1pt}
    \begin{subfigure}[b]{0.115\textwidth}
        \centering
        \includegraphics[width=\textwidth]{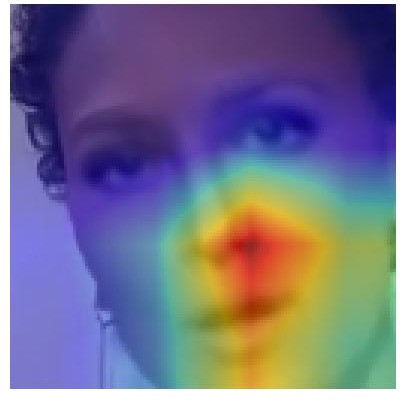}
        \caption*{(c) 60\% Sparse}
    \end{subfigure}
    \hspace{1pt}
    \begin{subfigure}[b]{0.115\textwidth}
        \centering
        \includegraphics[width=\textwidth]{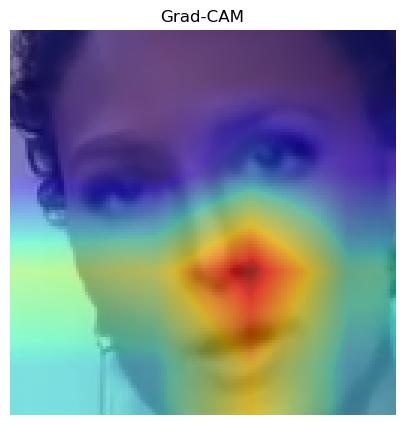}
        \caption*{(d) 80\% Sparse}
    \end{subfigure}

    \vspace{4pt}

    \begin{subfigure}[b]{0.115\textwidth}
        \centering
        \includegraphics[width=\textwidth]{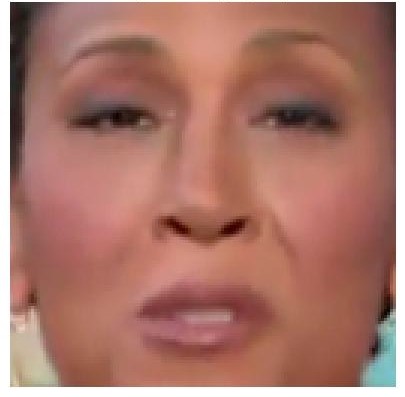}
        \caption*{(e) Real}
    \end{subfigure}
    \hspace{1pt}
    \begin{subfigure}[b]{0.115\textwidth}
        \centering
        \includegraphics[width=\textwidth]{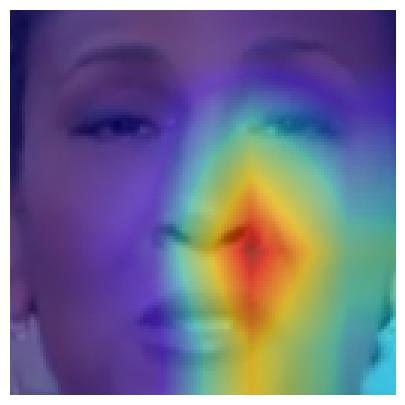}
        \caption*{(f) 0\% Sparse}
    \end{subfigure}
    \hspace{1pt}
    \begin{subfigure}[b]{0.115\textwidth}
        \centering
        \includegraphics[width=\textwidth]{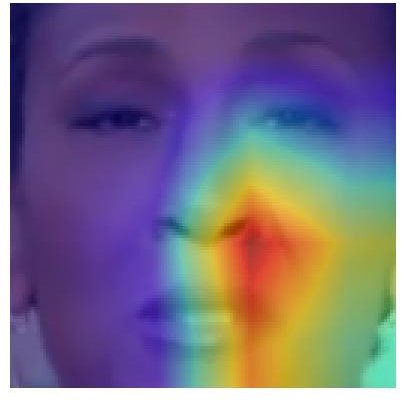}
        \caption*{(g) 60\% Sparse}
    \end{subfigure}
    \hspace{1pt}
    \begin{subfigure}[b]{0.115\textwidth}
        \centering
        \includegraphics[width=\textwidth]{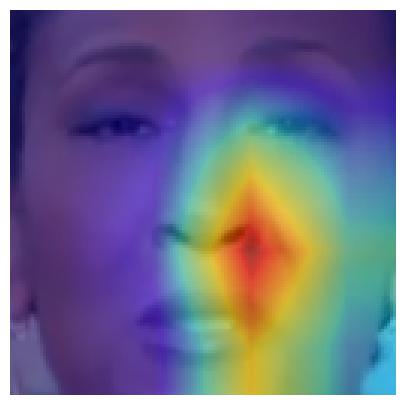}
        \caption*{(h) 80\% Sparse}
    \end{subfigure}

    \caption{
    Grad-CAM visualizations from a ResNet-18 model at varying sparsity levels for both fake (top row) and real (bottom row) images. As the sparsity increases, the highlighted regions evolve, indicating changes in feature attribution under pruning.
    }
    \label{fig:gradcam_comparison}
\end{figure*}

Based on the Grad-CAM visualization results shown in Fig.~\ref{fig:gradcam_comparison}, we can observe how network pruning affects the model's attention patterns when classifying real and fake facial images. The analysis compares three sparsity levels: 0\% (dense model), 60\%, and 80\% pruning. Interestingly, despite the significant reduction in model parameters, the Grad-CAM activation maps demonstrate remarkable consistency in attention regions across all sparsity levels. For both real and fake image classifications, the model primarily focuses on facial features in the right portion of the face, particularly concentrating on the cheek and eye regions. This consistent attention pattern suggests that the pruned models maintain their ability to identify relevant discriminative features even at high sparsity levels of 80\%. The heat maps show the highest activation (red regions) consistently positioned in similar areas across all sparsity levels, indicating that the essential feature extraction capabilities are preserved during pruning. This visualization provides evidence that the pruned models retain their ability to focus on semantically meaningful regions for deepfake detection, even when a significant portion of the network's parameters has been removed. One interesting observation is that, in both cases, the pruned model reduces its span of high-concentration areas and focuses only on the regions that appear to be important. Since pruned models cannot cover a wide area, this could explain their poor performance in extreme sparsity.
 
\begin{table}[tb]
\caption{The performance of subnetworks that are transferred from models trained on the larger OpenForensic dataset to those trained on the smaller Forensic++ dataset.}
\centering
\begin{tabular}{cccccc}
\toprule
\multirow{2}{*}{\textbf{Model}} & \multirow{2}{*}{\textbf{PR}} & \multirow{2}{*}{\textbf{Param (M)}} & \multicolumn{3}{c}{\textbf{Forensic++ Accuracy (\%)}} \\
\cmidrule(lr){4-6}
& & & \textbf{Base} & \textbf{IMP} & \textbf{OneShot} \\
\midrule
MesoNet & 0\% & 0.015 & 59.9 & - & - \\
Ours & 60\% & 0.006 & - & \textbf{61.2} & 59.8 \\
Ours & 80\% & \textbf{0.003} & - & 59.3 & 55.9 \\
\midrule
CNN-5 & 0\% & 0.300 & 76.2 & - & - \\
Ours & 60\% & 0.12 & - & \textbf{76.1} & 75.5 \\
Ours & 80\% & \textbf{0.06} & - & 74.5 & 73.3 \\
\midrule
ResNet-18 & 0\% & 11.700 & 85.3 & - & - \\
Ours & 60\% & 4.680 & - & \textbf{85.6} & 84.9 \\
Ours & 80\% & \textbf{2.34} & - & 83.7 & 83.3 \\
\bottomrule
\end{tabular}
\begin{tablenotes}
\small
\item \textbf{PR:} Pruning Rate — the percentage of weights removed from the model. For example, 60\% PR means 40\% of the original weights remain.
\end{tablenotes}
\label{tab:transfer_results}
\end{table}

\subsection{Transfer Ticket}

 Sparse subnetworks, identified through different pruning levels, demonstrate significant transferability across tasks, highlighting their utility in resource-constrained environments. Table \ref{tab:transfer_results} presents the performance of sparse subnetworks transferred from models trained on the larger OpenForensic dataset to models trained on the smaller Forensic++ dataset. Key observations include:

\textbf{(1)} Subnetworks pruned to 60\% sparsity consistently maintain or even improve accuracy compared to their dense counterparts. For instance, at 60\% sparsity, our method achieves 61.2\% accuracy with IMP and 59.8\% with one-shot pruning, retaining almost the dense MesoNet baseline (59.9\%). 

\textbf{(2)} Extremely sparse subnetworks (80\% sparsity) achieve competitive results despite their drastic reduction in parameters. Remarkably, with only 0.06M parameters, the 80\% sparse CNN-5 subnetwork retains almost similar accuracy, demonstrating the effectiveness of pruning in retaining critical features. 

\textbf{(3)} Transferability improves with higher-performing base models. For example, CNN-5 subnetworks transferred at 80\% sparsity achieve 76.1\% accuracy with IMP pruning, outperforming MesoNet in similar setups, particularly similar parameters. This highlights that the subnetworks effectively capture and adapt fundamental features from the source dataset.

\section{Limitations, Conclusions, and Future Work} 

Our study has three primary limitations: (1) Frame-level analysis from FaceForensics++ may not fully capture temporal dynamics in video-based deepfakes; (2) The pruning criteria are not specifically optimized for preserving manipulation artifacts; and (3) The computational overhead of IMP limits real-time applicability. 

Despite these limitations, our investigation of the Lottery Ticket Hypothesis in deepfake detection across MesoNet, CNN-5, ResNet-18, and XceptionNet on the OpenForensic and FaceForensics++ datasets demonstrates that deepfake detection networks contain winning tickets capable of maintaining robust performance even under extreme sparsity. Our iterative pruning approach consistently outperforms one-shot pruning, especially at higher sparsity levels. In particular, we observe that the effectiveness of pruning strategies varies between datasets: IMP-Global performs better on OpenForensic, while IMP-Local shows superiority on FaceForensics++, highlighting the importance of dataset-specific pruning approaches for optimal performance in deepfake detection. We fixed the pruning increment at 20\% per iteration. However, prior research\cite{frankle2020stabilizinglotterytickethypothesis,chen2022coarsening} suggests that smaller pruning steps (e.g., 5–10\%) may better preserve rare or class-specific features. Investigating the sensitivity of winning ticket discovery to pruning granularity remains a valuable future direction.

Future work should focus on (1) temporal-aware pruning strategies for video-based detection, (2) examining the relationship between network sparsity and robustness against adversarial attacks, and (3) exploring the transferability of winning tickets across deepfake generation methods. Efficient retraining strategies could also reduce the computational cost of identifying winning tickets, enabling real-time deepfake detection. Although we primarily compare our method against one-shot pruning, integrating comparisons with Single-shot Network Pruning (SNIP), Gradient Signal Preservation (GraSP), or layer-wise saliency pruning techniques \cite{lee2019snipsingleshotnetworkpruning,Wang2020Picking} is a valuable future direction. However, prior work \cite{frankle2018lottery,8485719} indicates that iterative pruning methods generally outperform one-shot criteria for sparse training in vision models.

\bibliographystyle{IEEEtran}
\balance
\bibliography{ref}

\end{document}